# SFS-A68: a dataset for the segmentation of space functions in apartment buildings

Ziaee A., Suter G.
Design Computing Group, TU Wien, Vienna, Austria
amir.ziaee@tuwien.ac.at

**Abstract.** Analyzing building models for usable area, building safety, or energy analysis requires function classification data of spaces and related objects. Automated space function classification is desirable to reduce input model preparation effort and errors. Existing space function classifiers use space feature vectors or space connectivity graphs as input. The application of deep learning (DL) image segmentation methods to space function classification has not been studied. As an initial step towards addressing this gap, we present a dataset, SFS-A68, that consists of input and ground truth images generated from 68 digital 3D models of space layouts of apartment buildings. The dataset is suitable for developing DL models for space function segmentation. We use the dataset to train and evaluate an experimental space function segmentation network based on transfer learning and training from scratch. Test results confirm the applicability of DL image segmentation for space function classification.

## 1. Introduction

Building information modeling (BIM) authoring systems support detailed analysis of usable area, building safety, or energy analysis (e.g., Autodesk, 2021). Accurate analysis requires function classification data for spaces and related objects. Usually, these data need to be manually entered by end-users. They may also be lost in data exchanges between BIM authoring software (Lai et al., 2019). This is problematic because it can lead to input errors and inaccurate analysis. Furthermore, input preparation is time-consuming, particularly for large and functionally diverse buildings. Automating the space function classification task is desirable because it can help reduce input preparation effort and input errors and improve the accuracy of building analysis results.

To automate classification tasks in BIM, researchers have explored the application of supervised Machine Learning (ML) and Deep Learning (DL) methods (Section 2). Segmentation, as the most well-known capability of supervised ML and DL, can classify a variety of object classes at the pixel level of an image (Long et al., 2015). While DL segmentation methods have been developed to segment building and furnishing elements in point clouds (Ma, Czerniawski and Leite, 2020; Emunds et al., 2021), their application to space function classification has not been studied. A fundamental challenge is to develop suitable datasets to train and evaluate segmentation models. To address this issue, we present a dataset, SFS-A68, that consists of input and ground truth images generated from 68 digital 3D models of space layouts of apartment buildings (Ziaee, Suter and Barada, 2022). We use the dataset to train and evaluate an experimental space function segmentation network based on transfer learning (Tan et al., 2018) and training from scratch. We follow a general ML and DL workflow (Figure 1) consisting of problem formulation, data collection, pre-processing, construction, validation, and model deployment (Eisler and Meyer, 2020).

First, we define space function classification as a semantic segmentation problem. We choose space functions in apartment buildings as an application domain. Apartments are typically diverse in space areas, functions, and accessibility (Heckmann and Schneider, 2011), making them suitable to explore the application potential of segmentation models. Next, we describe the data collection workflow which we have adopted to develop our dataset. The workflow involves selecting buildings, interpreting their floor plans, modeling space layouts in BIM authoring or CAD systems, and data quality verification. Layouts are pre-processed according to the input requirements for space function segmentation. We describe how vector images of layout models are automatically converted to raster images. Layout elements are colored consistently by their class in these images. We have used the dataset to train and validate a space function segmentation network based on transfer learning and training from scratch. We present initial results obtained from testing the network's capability to predict space functions. Deployment of the space function segmentation network in a BIM analysis application is beyond the scope of this study. It would require the post-processing of prediction images. In this step, a predicted class or a list of predicted classes would need to be extracted for each space from all pixels in a prediction image that carry class prediction data for that space. Information about predicted classes would be passed to the BIM analysis application to modify space labels in the building model accordingly.

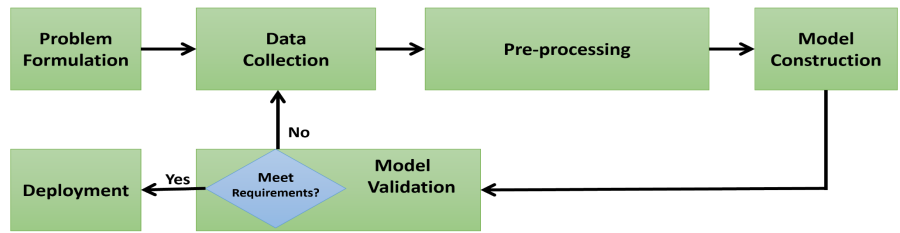

Figure 1: Machine Learning and Deep Learning workflow (Eisler and Meyer, 2020).

## 2. Related work

Classification methods for BIM can be categorized into rule-based, ML-based, and DL-based methods. Rule-based classification involves rule sets to encapsulate domain knowledge. For example, Belsky et al. (2016) introduced a rule language to classify objects based on their properties and spatial relationships. Bloch and Sacks (2018) developed rules on single or pairwise room features to classify space functions in apartment buildings. Examples for features include room area, accessibility, or the number of windows in a room. Suter (2022) developed space ontologies encoded in the Ontology Web Language (OWL) and used a general-purpose OWL semantic reasoner to infer space functions of given space layouts. Rule-based methods are limited because knowledge needs to be encoded explicitly. Developing robust and non-contradictory rules is a challenging task that requires domain and programming expertise. For space function classification, the accuracy of rule-based methods may be lower than ML and DL methods (Bloch and Sacks, 2018).

By contrast, supervised ML and DL are data-driven methods. They use a set of inputs with their associated labels or ground truths to train a model that can be used to classify new inputs. A key difference between the two methods is that DL has the capability to extract features automatically. De las Heras et al. (2014) proposed a two-step pipeline to interpret floor plan images. In the first step, walls, doors, and windows are segmented using supervised ML. In the second step, a wall junction graph is derived, and room regions are identified as cycles in the graph. Bloch and Sacks (2018) applied supervised ML with iteration to classify space functions in apartments. Their artificial neural network (ANN) classifier uses features of individual spaces, including floor area and the number of doors, and space connectivity features as input. The ANN was trained to classify 15 space functions for a dataset of 150 spaces from 32 similar apartments.

Applications of DL methods in BIM include semantic segmentation of building interiors. Ma, Czerniawski and Leite (2020) created DL models to segment building and furnishing element classes in point clouds of non-residential interior spaces. Synthetic point clouds were generated from BIM models using BIM authoring and CAD software. Leonhardt et al. (2020) converted geometries of objects in room scenes to synthetic point clouds to train a DL neural network for 3D object classification and segmentation. IFCNet is a dataset consisting of approximately 19'000 instances of 65 IFC classes (Emunds et al., 2021). Instance images were created from multiple viewpoints and passed as input to a Multi-View Convolutional Neural Network (MVCNN) to learn shape descriptors. Wang, Sacks and Yeung (2022) developed a Graph Neural Network (GNN) model that classifies eight apartment space functions. A dataset was created that consists of spatial connectivity graphs for 224 apartment layouts from three countries.

Existing work indicates a significant potential for ML and DL methods to address classification needs in BIM in general and space function classification in particular. Existing space function classifiers use space feature vectors or space connectivity graphs as input (Bloch and Sacks, 2018; Wang, Sacks and Yeung, 2022). However, the applicability of DL image segmentation methods to space function classification has not been studied. This gap appears significant because DL image segmentation methods have been applied successfully in many domains (Minaee et al., 2020). Our SFS-A68 dataset provides a basis for developing novel space function classifiers that use such methods.

## 3. Problem formulation

We aim to segment space function classes in images of space layout models of apartment buildings. Examples for space function classes are 'LivingRoom', 'Kitchen', 'Loggia' and 'Elevator'. There are two variants of the problem: semantic and instance space function segmentation. The task of semantic space function segmentation is to assign each pixel in a space layout image to a space function class. Instance segmentation assigns each space pixel to a space instance of a space function class. In this study, we focus on semantic space function segmentation. We aim to develop DL image segmentation networks for semantic space function segmentation (Figure 2).

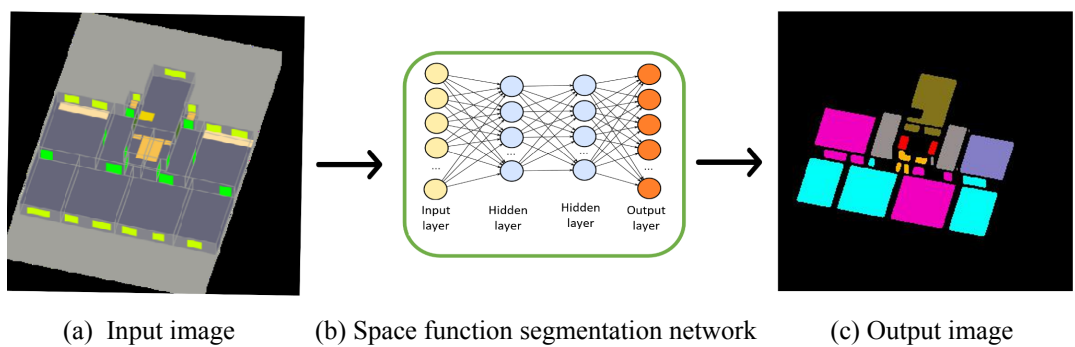

(a) Input image     (b) Space function segmentation network     (c) Output image

Figure 2: Space function segmentation network.

To train a space function segmentation network, we need input images of space layouts with dimensions H x W x 3 as uniformly scaling two-dimensional RGB images and corresponding ground truth images with dimensions H x W x N, where N corresponds to the number of space function classes. Each layout element in an input image is colored according to a unique class color (Table 1, left column). Space element classes may be helpful as metadata to determine space function classes. For example, a 'Bathroom' may contain a 'SanitaryElement'.

Each pixel in a ground truth image is assigned to a space function class. Space function classes in apartment buildings that are classified by the space function segmentation network are shown in Table 1, right column. Circulation spaces, such as stairways or elevators, are classified because we aim to classify complete floors rather than individual apartments. We have identified 22 space function classes for the apartment buildings in our dataset. Since existing space classification systems, such as UniClass and OmniClass (NBS, 2020; CSI, 2010), currently do not cover all required classes, we defined a class hierarchy that meets our needs.

Table 1: Class hierarchies.

| Layout element classes | | Space function classes | | |
|---|---|---|---|---|
| **Name** | **Count** | **Name** | **Count [-]** | **[%]** |
| Space | | Space | | |
| InternalSpace | 2'296 | **ResidentialSpace** | | |
| ExternalSpace | 223 | CommunalSpace | | |
| SpaceElement | | DiningRoom | 3 | 0.1 |
| **SpaceContainedElement** | | FamilyRoom | 6 | 0.2 |
| CirculationElement | | LivingRoom | 275 | 10.9 |
| FlightOfStairs | | PrivateSpace | | |
| Landing | | Bedroom | 495 | 19.7 |
| FurnishingElement | | MasterBedroom | 23 | 0.9 |
| KitchenElement | | BoxRoom | 2 | 0.1 |
| SanitaryElemen | | HomeOffice | 8 | 0.3 |
| EquipmentElement | | **ServiceSpace** | | |
| HomeAppliance | | Shaft | 401 | 15.9 |
| TextileCareAppliance | | StorageRoom | 83 | 3.3 |
| **SpaceEnclosingElement** | | WalkInCloset | 2 | 0.1 |
| Opening | | SanitarySpace | | |
| Partition | | Bathroom | 301 | 11.9 |
| Window | | Toilet | 150 | 6.0 |
| Door | | Kitchen | 25 | 1.0 |
| RegularDoor | | LaundryRoom | 117 | 4.6 |
| UnitDoor | | **CirculationSpace** | | |
| ElevatorDoor | | VerticalCirculationSpace | | |
| | | Elevator | 84 | 3.3 |
| | | Stairway | 70 | 2.8 |
| | | HorizontalCirculationSpace | | |
| | | Entrance | 75 | 3.0 |
| | | Hallway | 11 | 0.4 |
| | | MainHallway | 19 | 0.8 |
| | | InternalHallway | 145 | 5.8 |
| | | **External** | | |
| | | AccessBalcony | 19 | 0.8 |
| | | Loggia | 108 | 4.3 |
| | | Air | 96 | 3.8 |

## 4. Data collection

The SFS-A68 dataset is derived from source data comprising 68 digital 3D models of space layouts of apartment buildings designed or built in the years 1952-2019. Master of Architecture students at TU Wien created the models for an Architectural Morphology course taught by the second author. Each layout model represents a complete or partial floor plan. In total, source data cover 275 apartments of varying types (Table 2). 52% of apartments are 2-Room or 3-Room apartments.

Table 2. Apartment types in the SFS-A68 dataset.

| Type | Count [-] | [%] |
|---|---|---|
| Apartment | 275 | 100 |
| 1-Room Apartment | 17 | 6 |
| 2-Room Apartment | 65 | 24 |
| 3-Room Apartment | 78 | 28 |
| 4-Room Apartment | 53 | 19 |
| 5-Room Apartment | 49 | 18 |
| >5-Room Apartment | 13 | 5 |

Students selected floor plans from a variety of sources, including books and professional journals on apartment design and architecture websites (Heckmann and Schneider, 2011; Plataforma Networks, 2022). 78% of floor plans are from buildings in Austria, Germany, and Switzerland, and 22% from 10 countries in Europe, the USA, and Southern America, respectively. Students created space layout models in BIM authoring (2021) or CAD (2015-2020) software by following a four-step workflow.

In the first step, an apartment building is selected. The building must be well documented. At a minimum, floor plans must be available either in raster image or vector drawing formats, and their scales must be known. Floor plans and related data are considered ground truth.

In the second step, floor plans of a selected apartment building are interpreted. Ambiguous features in floor plans are identified and resolved. For example, the physical separation between a living room and a hallway or a kitchen may be partial or non-existent.

In the third step, a layout model is created using BIM authoring or CAD software for a selected floor plan. The model is exported as an IFC file and imported into the Space Modeling System (SMS; Suter, 2022). SMS extracts space data from the IFC file and simplifies the shapes of space elements. For example, window shapes are converted from a multiple Brep to a single face representation. A label editor in SMS is used to semi-automatically label layout elements according to Table 1. A detailed description of this step is given in (Suter, 2022).

In the fourth step, the data quality of the layout model and labels is verified. Typical inconsistencies include missing doors or layout elements with inaccurate labels. The data quality of each layout model in the SFS-A68 dataset was verified independently against ground truth floor plans by a teaching assistant and the second author, who both have an architectural background.

## 5. Pre-processing

3D space layout models are converted to SVG images where layout element regions are colored by layout element class. Next, the colored SVGs are converted to PNG raster images that can be used as input or ground truth to train space function segmentation networks. Each image pixel value is normalized to make non-zero gradients less frequent during training and help networks learn faster.

### 5.1 Conversion to SVG

SMS converts a 3D space layout model to two Scalable Vector Graphics (SVG) images whose elements are colored by their class. SVG is a W3C standard for 2D graphics based on XML syntax (W3C, 2018). It supports vector objects, such as paths (regions), raster images, and text. A benefit of the SVG format for the pre-processing workflow is that the appearance of image elements can be defined in linked Cascading Style Sheet (CSS) style sheets. SMS projects 3D layout elements on the XY plane of the layout model's local coordinate system. Layout element faces are converted to regions. Each region has a class attribute whose value is populated with the class of its layout element. A style is defined for each layout element class in a CSS style sheet. It has properties for stroke thickness and fill color. Each layout element class has a unique fill color (Table 1). As a result, SVG images of layout models generated by SMS are colored consistently. A single CSS style sheet is used to color all SVG images for input and ground truth. Each space region in an input image is colored as 'InternalSpace' or 'ExternalSpace' while a region for a ground truth image is colored by its function class. Space elements are colored according to their classes in each input image.

### 5.2 Conversion to PNG

SVG images are converted to raster images. The PNG raster image format is chosen for this study. We use the Wand library (ImageMagick Studio, 2021) to convert an SVG image for input or ground truth to a corresponding PNG image. The shape of an input image returned by the procedure is Height x Width x [R, G, B]. For ground truth, all space regions are iteratively extracted for each space function class from the SVG image, and then they are converted to a PNG image with shape Height x Width x Number of Space Function Classes.

### 5.3 Normalization

Normalization is widely used in the data pre-processing steps of DL models to decrease a model's training time and prevent model weights from increasing during training such that networks converge faster. Scaling of pixel channels from the [0, 255] range to ranges [0, 1] and [-1, 1] are two commonly used approaches to ensure that each pixel has a similar data distribution (Huang et al., 2020). The first scaling approach is known as normalization and is applied to a ground truth image using Otsu's thresholding function in the OpenCV library after Gaussian filtering (Bradski, 2000). Subsequently, binary masks are created for the ground truth image. In a binary mask, 0 represents a background pixel, and 1 represents a foreground space function class pixel. The second scaling approach is known as standardization. It shifts the data distribution to have a mean of zero and a standard deviation of one unit. It is performed per pixel as shown in Equation 1:

$$X' = (X / \mu) - \sigma \qquad (1)$$

where $X'$ is the normalized pixel, $\mu$ and $\sigma$ are the mean and the standard deviation over all channels, respectively (Huang et al., 2020).

### 6. Model construction

Applying the pre-processing workflow to each layout results in a dataset comprised of 68 standardized input images with their corresponding normalized ground truth images. Since the dataset used to train a model must be different from the one used to evaluate its performance, the SFS-A8 dataset is split into a training and a testing dataset with a ratio of 80% (54 layouts) to 20% (14 layouts). In this study, we do not perform hyperparameter fine-tuning, which requires a validation dataset. Therefore, the training dataset is used to train the space function segmentation network and the test dataset to evaluate it.

The U-Net (Siddique et al., 2021) architecture is chosen to segment space function classes. It is a fully convolutional network consisting of an encoder-decoder with a skip connector from the encoder to the decoder. The encoder's task is to downsample and gradually reduce the spatial dimension for feature extraction, and the decoder's task is to restore image details and spatial dimension by up-sampling. We decided to use Training from Scratch (TfS U-Net) and Transfer Learning (TL U-Net) as two approaches for training the U-Net. Transfer learning (Tan et al., 2018) is useful when available training data are limited, thereby saving resources and improving the accuracy and training time of new models. The typical approach to perform transfer learning with a segmentation network is to use a pre-trained classification model as an encoder and build a decoder on top of it. The decoder is trained first, followed by the encoder, and finally, both are trained on a small data set. Transfer learning the space function segmentation model is done as follows:

1. **Select a pre-trained classification model:** VGG16 (Simonyan and Zisserman, 2014), which was trained on a large dataset called ImageNet (Deng et al., 2009), was chosen as the U-net encoder because it has fewer convolutional layers between max-pooling layers, which more closely matches the original U-net structure and therefore training is faster.

2. **Create a decoder:** A decoder is created on top of the encoder by adding three blocks, where each block is consists of skip connection, convolution, batch normalization, LeakyRelu (Xu et al., 2015), deconvolution, batch normalization, and LeakyRelu. Skip connections were used to connect the encoder layers to the decoder blocks. The output number of the decoder's last layer corresponds to the number of space function classes in the dataset.

3. **Train the model:** A strategy was developed to train the decoder and encoder. The strategy consists of three steps:
   1. Train the decoder with a learning rate of 0.001 and an epoch number of 100, while the pre-trained U2-Net encoder layers are frozen.
   2. Unfreeze all the encoder layers and train them with a learning rate of 0.0001 and an epoch number of 100 while the decoder layers are frozen.
   3. Unfreeze both the decoder and encoder and train them simultaneously with a learning rate of 0.00009 and an epoch number of 100.

To train U-net from scratch, we used the developed training strategy in reverse order, starting with the third step with a learning rate of 0.001, then the second step with a learning rate of 0.0001, and finally the first step with a learning rate of 0.0001. Since the training dataset is small and different from the dataset of the pre-trained model, in each training step, input and ground truth images are augmented by resizing to H+30 × W+30 as well as random rotation and flipping vertically and horizontally, followed by cropping to H × W. During training of the networks, an Adam optimizer (Zhang, 2018) is used to minimize the binary cross-entropy loss (Equation 2):

$$L = -\frac{1}{N}\sum_{i=1}^{N} y_i * log\,\widehat{y_i} + \left(1 - y_i\right) * log\left(1 - \widehat{y_i}\right) \qquad (2)$$

where $y_i$ is the corresponding target value of the i-th scalar value in the network output ($\widehat{y_i}$), and N is the number of scalar values in the model output.

## 7. Model validation

The outputs of the space function segmentation models for the test dataset are evaluated by Intersection over Union (IoU) and total error metrics (Long et al., 2015). IoU quantifies the percent overlap between ground truth and prediction output based on Equation 3:

$$IoU = \frac{TP}{TP + FP + FN} \qquad (3)$$

where TP is the number of true-positive pixels, FP is the number of false-positive pixels, and FN is the number of false-negative pixels. Total error computes the number of all incorrect pixels divided by the total number of pixels based on Equation 4:

$$Total\ Error = \frac{FN + FP}{FN + FP + TN + TP} \qquad (4)$$

where TN is the number of true-negative pixels. Evaluation results for the test dataset are given in Table 3. IoU values are computed only for classes that are present in a test layout's ground truth. For example, 'FamilyRoom' and 'WalkInCloset' are not present in any test layout, and therefore no IoU values are reported for these classes. Examples from the test dataset are shown in Figure 5.

Table 3: Intersection over Union (IoU) and Total Error for the test dataset. IoU is computed only for classes that are present in a test layout's ground truth.

| Class | Number of test layouts with class | IoU | | Total Error | |
|---|---|---|---|---|---|
| | | TfS U-Net | TL U-Net | TfS U-Net | TL U-Net |
| DiningRoom | 1 | 0.00 | 0.33 | 0.000 | 0.000 |
| FamilyRoom | 0 | n/a | n/a | 0.000 | 0.000 |
| LivingRoom | 14 | 0.35 | 0.94 | 0.057 | 0.005 |
| Bedroom | 13 | 0.00 | 0.90 | 0.044 | 0.004 |
| MasterBedRoom | 5 | 0.00 | 0.60 | 0.007 | 0.002 |
| BoxRoom | 1 | 0.00 | 0.89 | 0.001 | 0.000 |
| HomeOffice | 1 | 0.00 | 0.75 | 0.001 | 0.000 |
| Shaft | 12 | 0.00 | 0.05 | 0.001 | 0.001 |
| StorageRoom | 6 | 0.00 | 0.25 | 0.001 | 0.001 |
| WalkInCloset | 0 | n/a | n/a | 0.000 | 0.000 |
| BathRoom | 14 | 0.00 | 0.71 | 0.009 | 0.003 |
| Toilet | 9 | 0.00 | 0.19 | 0.001 | 0.001 |
| Kitchen | 8 | 0.00 | 0.82 | 0.007 | 0.001 |
| LaundryRoom | 4 | 0.00 | 0.09 | 0.001 | 0.001 |
| Elevator | 12 | 0.00 | 0.69 | 0.002 | 0.001 |
| Stairway | 11 | 0.00 | 0.82 | 0.009 | 0.001 |
| Entrance | 6 | 0.00 | 0.59 | 0.005 | 0.002 |
| Hallway | 1 | 0.00 | 0.53 | 0.000 | 0.000 |
| MainHallway | 3 | 0.00 | 0.70 | 0.002 | 0.001 |
| InternalHallway | 6 | 0.00 | 0.44 | 0.005 | 0.002 |
| AccessBalcony | 3 | 0.00 | 0.90 | 0.006 | 0.001 |
| Loggia | 3 | 0.00 | 0.90 | 0.002 | 0.000 |

## 8. Discussion

The results show that the TL U-Net model outperforms the TfS U-Net model for IoU and Total Error metrics for all space function classes. The TfS U-Net model only predicts the 'LivingRoom' class. It does not make predictions for the other classes (Figure 5d). The training dataset appears too small for training from scratch.

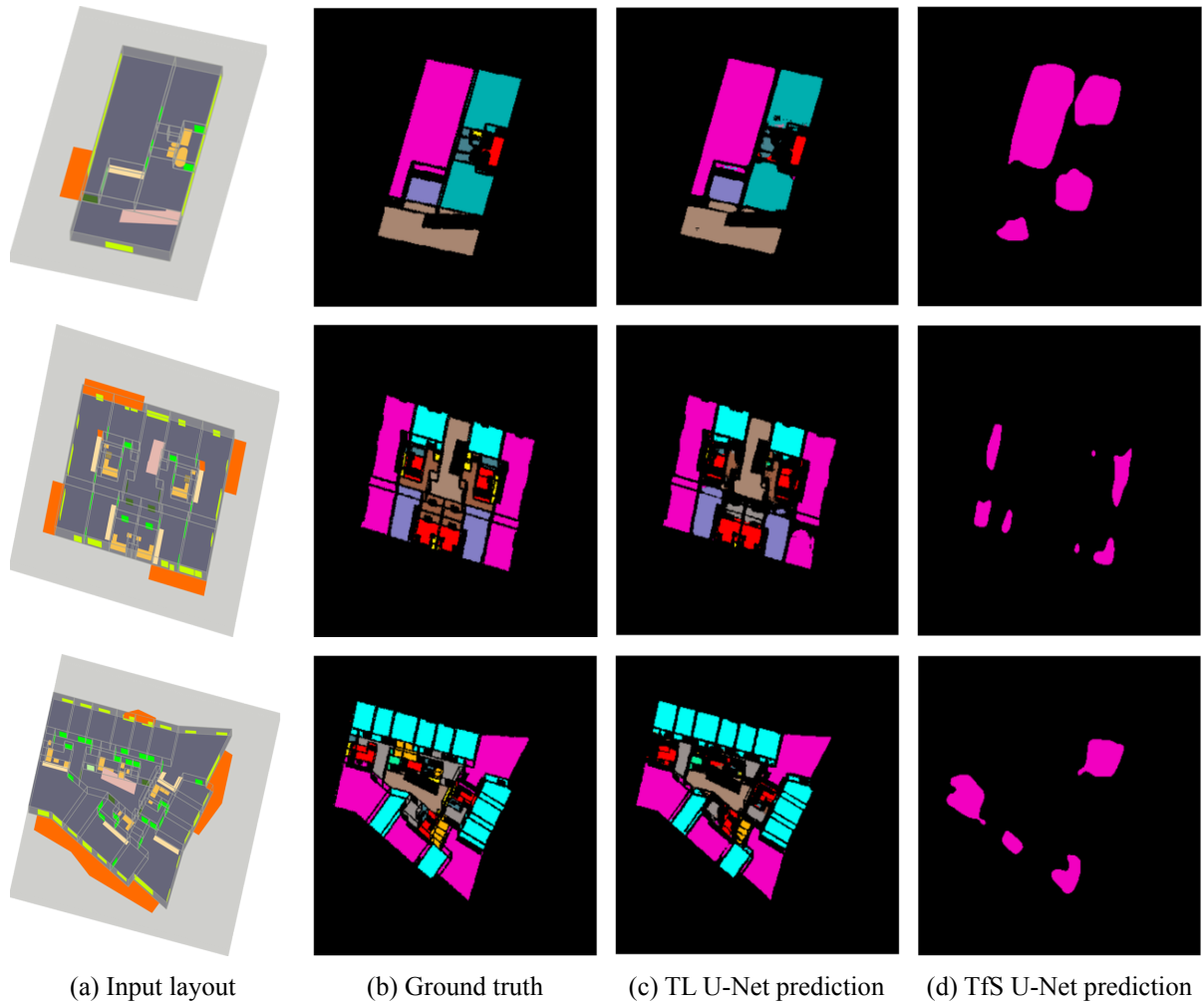

(a) Input layout          (b) Ground truth          (c) TL U-Net prediction          (d) TfS U-Net prediction

Figure 5: Test dataset examples.

The TL U-Net model makes predictions for all classes, but significant performance differences exist. IoU values are highest (IoU>=0.9) for 'LivingRoom', 'Bedroom', and external spaces. They are the lowest (IoU<=0.25) for small service spaces. This result is qualitatively similar to an existing space function classification model (Wong, Sacks, and Yeung, 2022), even though the number of predicted classes and the DL method used differ from the TfS U-Net model. IoU values do not appear to depend on the frequency of classes in the SFS-A68 dataset. For example, only 5.1% of spaces in the SFS-A68 dataset are 'Loggia' or 'AccessBalcony' spaces (Table 1), but the TL U-Net model is better at predicting these classes than 'Shaft' spaces, which make up 15.9% of spaces. As an explanation, 'Loggia' and 'AccessBalcony' spaces are larger, and they (or their context) may have more salient features than 'Shaft' spaces.

The SFS-A68 dataset currently has two main limitations. First, layout images are created by projection from a single, fixed viewpoint only. Instead, suppose the dataset included images taken from multiple viewpoints of the space layout models, similar to Edmunds et al. (2021). In that case, a space function segmentation network may more easily detect certain features, such as doors or windows. Second, as 78% of buildings are located in the German-speaking region of Europe, the dataset is not well balanced regarding geography, climate, culture, or building codes. We aim to address these limitations in future versions of the dataset.

## 9. Conclusion

We presented a dataset for space function segmentation in apartment buildings. We plan to extend the dataset to further increase the diversity of space layouts, particularly regarding location. Since SMS is a general-purpose layout modeling system, it could be used to create similar datasets for other building types, such as office buildings or schools.

We used the dataset to develop segmentation models based on transfer learning and training from scratch. In future work, we plan to compare the performance of different DL model architectures, including hyperparameter fine-tuning. Moreover, the influence of space elements or viewpoints on model performance needs to be investigated. Finally, we will explore how predictions of a space function segmentation network can be deployed in BIM analysis applications.

## Acknowledgments

The authors gratefully acknowledge support by Grant Austrian Science Fund (FWF): I 5171-N, Mihael Barada, and participants in course '259.428-2021S Architectural Morphology' at TU Wien for data collection.